\documentclass[conference]{IEEEtran}
\IEEEoverridecommandlockouts
\usepackage{cite}
\usepackage{amsmath,amssymb,amsfonts}
\usepackage{algorithmic}
\usepackage{graphicx}
\usepackage{textcomp}
\usepackage{xcolor}

\usepackage{booktabs}
\usepackage{xcolor}

\def\BibTeX{{\rm B\kern-.05em{\sc i\kern-.025em b}\kern-.08em
    T\kern-.1667em\lower.7ex\hbox{E}\kern-.125emX}}

\begin{document}

\title{Vision Transformer Computation and Resilience \\for Dynamic Inference
\thanks{$^1$Kavya Sreedhar was supported by an internship at NVIDIA and by a graduate fellowship award as a Knight-Hennessy Scholar at Stanford University.}
}

\author{
\IEEEauthorblockN{Kavya Sreedhar$^1$}
\IEEEauthorblockA{
% \textit{Department of Electrical Engineering} \\
\textit{Stanford University}\\
Stanford, USA \\
skavya@stanford.edu}
\and
\IEEEauthorblockN{Jason Clemons}
\IEEEauthorblockA{
% \textit{Architecture Research Group} \\
\textit{NVIDIA}\\
Austin, USA \\
jclemons@nvidia.com}
\and
\IEEEauthorblockN{Rangharajan Venkatesan}
\IEEEauthorblockA{
% \textit{ASIC \& VLSI Research Group} \\
\textit{NVIDIA}\\
Santa Clara, USA \\
rangharajanv@nvidia.com}
\and
\IEEEauthorblockN{Stephen W. Keckler}
\IEEEauthorblockA{
% \textit{Architecture Research Group} \\
\textit{NVIDIA}\\
Austin, USA \\
skeckler@nvidia.com}
\and
\IEEEauthorblockN{Mark Horowitz}
\IEEEauthorblockA{
% \textit{Departments of Electrical Engineering and Computer Science} \\
\textit{Stanford University}\\
Stanford, USA \\
horowitz@ee.stanford.edu}
}

\maketitle

\begin{abstract}

State-of-the-art deep learning models for computer vision tasks are based on the transformer architecture and often deployed in real-time applications. In this scenario, the resources available for every inference can vary, so it is useful to be able to dynamically adapt execution to trade accuracy for efficiency. To create dynamic models, we leverage the resilience of vision transformers to pruning and switch between different scaled versions of a model. Surprisingly, we find that most FLOPs are generated by convolutions, not attention. These relative FLOP counts are not a good predictor of GPU performance since GPUs have special optimizations for convolutions. Some models are fairly resilient and their model execution can be adapted without retraining, while all models achieve better accuracy with retraining alternative execution paths. These insights mean that we can leverage CNN accelerators and these alternative execution paths to enable efficient and dynamic vision transformer inference. Our analysis shows that leveraging this type of dynamic execution can lead to saving 28\% of energy with a 1.4\% accuracy drop for SegFormer (63 GFLOPs), with no additional training, and 53\% of energy for ResNet-50 (4 GFLOPs) with a 3.3\% accuracy drop by switching between pretrained Once-For-All models.

\end{abstract}
\begin{IEEEkeywords}
vision transformers, dynamic inference, model resilience, pruning
\end{IEEEkeywords}

\vspace{-0.15in}

\section{Introduction}
\label{sec:introduction}

Deep learning models in computer vision have shifted from convolutional neural networks (CNNs)~\cite{he2017mask, girshick2015fast, ren2015faster, girshick2014rich, redmon2016you, liu2016ssd,  zheng2021rethinking, long2015fully, he2016deep, simonyan2014very, tan2019efficientnet, szegedy2016rethinking} to transformers~\cite{carion2020end, liu2022dabdetr, wang2022anchor, meng2021-CondDETR, xie2021segformer, liu2021swin, openai2023gpt4vision, dosovitskiy2020image, touvron2021training, kirillov2023segment, hatamizadeh2023global, chen2021crossvit, zhu2020deformable, dai2021dynamic, zheng2020end, yao2021efficient, roh2021sparse, strudel2021segmenter, zhang2022dino, zong2023detrs, zhou2022detecting} for higher model accuracy. The transformer architecture~\cite{vaswani2017attention} uses attention to understand global image contexts and effectively capture spatial information. It also underlies general-purpose backbones and foundation models for language~\cite{devlin2018bert, radford2018improving, radford2019language, brown2020language, achiam2023gpt, raffel2020exploring} and vision~\cite{liu2021swin, openai2023gpt4vision, dosovitskiy2020image, touvron2021training, kirillov2023segment} tasks. For example, in 2021, Microsoft introduced Swin Transformer~\cite{liu2021swin}, which has since been adopted as a backbone for various vision tasks~\cite{zhang2022dino, zong2023detrs, zhou2022detecting}, while Meta recently released the Segment Anything Model~\cite{kirillov2023segment} as a foundation model for segmentation tasks.

These models can be computationally expensive, requiring millions of parameters and billions of floating point operations (FLOPs)~\cite{khan2022transformers}. It is well-known that increasing the number of parameters and FLOPs in a model can lead to better model accuracy, as seen by recent trends in larger and more accurate GPT models, as an example~\cite{radford2018improving, radford2019language, brown2020language, achiam2023gpt, openai2023gpt4vision}.
Furthermore, these models typically have a fixed execution path and assume that their needed computational resources will be available when they are run.

% Vision transformers can be computationally expensive, requiring millions of parameters and billions of floating point operations (FLOPs)~\cite{khan2022transformers}, and typically have a fixed execution path. 

In contrast, real-time systems for applications such as autonomous driving~\cite{feng2020deep, siam2018comparative} and video conferencing~\cite{zhou2023survey} have limited hardware resources and dynamic system loads that change as the surrounding environment changes~\cite{sun2022resource, kim2023aurora, clemons2023leaf, wang2020dual}. Choosing a static model that matches the worst-case resource utilization would leave performance on the table when more resources are available in this scenario. As a result, these systems need to leverage dynamic vision transformers.

\begin{figure}[t]

  \centering

  \includegraphics[width=1\linewidth]{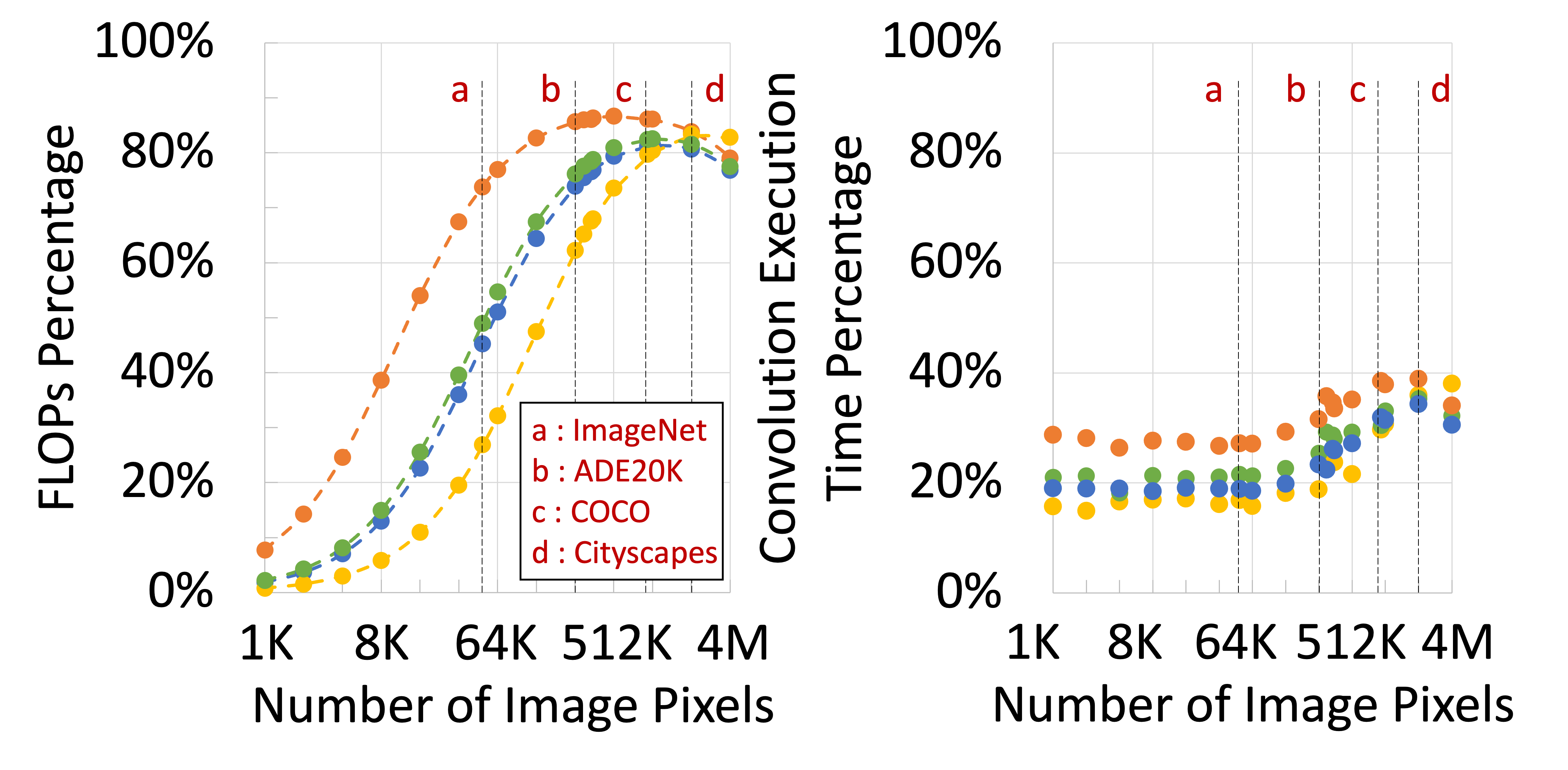}

  \vspace{-0.1in}

  \caption{FLOPs and NVIDIA RTX A5000 GPU execution time in convolutions (dots) and ResNet-50 backbone (dashed lines) for inference with DETR~\cite{carion2020end} (orange), Conditional DETR~\cite{meng2021-CondDETR} (green), DAB DETR~\cite{liu2022dabdetr} (blue), and Anchor DETR~\cite{wang2022anchor} (yellow). For larger image sizes, convolutions dominate FLOPs but not GPU execution time.}
  
  \label{figure_detr_flops}
  
% \vspace{-0.175in}
 
\end{figure}

Most prior work on dynamic models shortens the model execution based on complexity of classifying the input. These approaches exit early and remove the computation of later layers when internal predictions have already become stable~\cite{xin2020deebert, kaya2019shallow, hu2020triple, graves2016adaptive, liu2020fastbert, figurnov2017spatially, han2021dynamic, dehghani2018universal, teerapittayanon2016branchynet, zhou2020bert, tambe2021edgebert, wang2018skipnet, wang2020dual}. While this prior work reduces average latency and energy for ``easier'' inputs, it does not ensure that model execution meets a given dynamic resource constraint.

Some work has addressed this challenge by trading accuracy for efficiency in order to adapt the cost of inference to not exceed an input resource constraint~\cite{clemons2023leaf, wang2020dual}, but this prior work focuses on CNNs and BERT. In this paper, we extend that work for vision transformers. We build upon work that scales static model architectures and prunes redundant computation to achieve different levels of model computation and accuracy~\cite{carion2020end, jastrzkebski2017residual, michel2019sixteen, voita2019analyzing, cai2019once}. For example, Once-For-All (OFA) has developed efficient training techniques that produce many competitive subnetworks after training one model~\cite{cai2019once}.

% for background section
% \begin{figure*}[t]

%   \centering
  
%   \includegraphics[width=1\linewidth]{isca2023-latex-template/images/placeholder_star.png}
%   \caption{SegFormer and Swin Models}
  
%   \label{fig_segformer_swin_models}
  
%   \vspace{-0.1in}
 
% \end{figure*}

\begin{figure*}[t]

  \centering
  
  \includegraphics[width=1\linewidth]{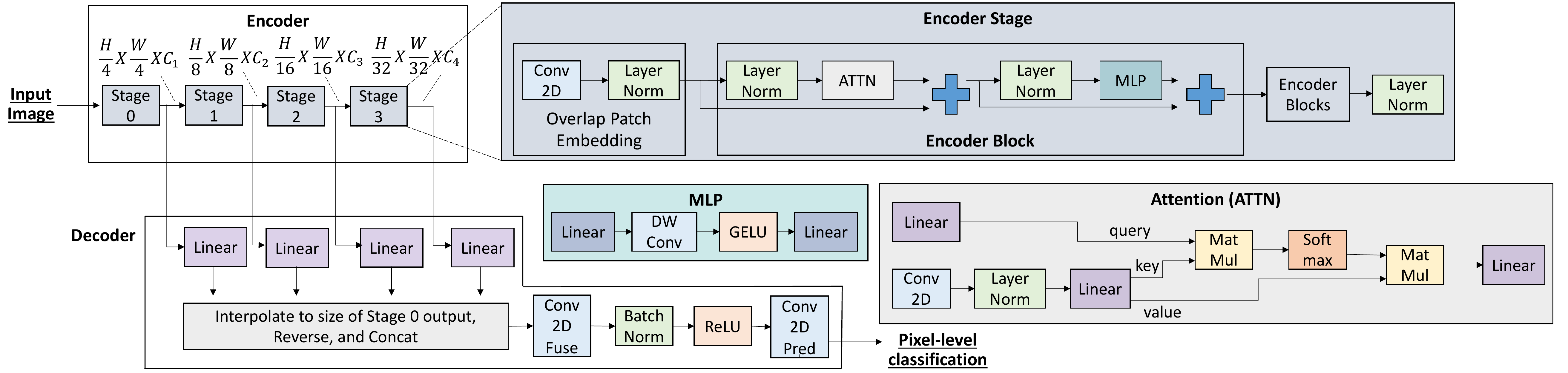}
  \caption{Layers in SegFormer~\cite{xie2021segformer} model. The Swin Transformer~\cite{liu2021swin} model follows the same high-level structure, with a more optimized attention module and the UPerNet decoder head~\cite{xiao2018unified}. The UPerNet decoder has a layer similar to SegFormer's \textit{Conv2DFuse}, which we refer to as \textit{fpn\_bottleneck\_Conv2D}.}
  
  \label{figure_segformer_swin_models}
  
% \vspace{-0.15in}
 
\end{figure*}

We examine resource-dependent dynamic (RDD) inference~\cite{clemons2023leaf, wang2020dual} for vision transformers by profiling this class of applications and identifying lower-cost execution paths. When examining the full application pipeline for state-of-the-art vision transformers~\cite{carion2020end, liu2022dabdetr, wang2022anchor, meng2021-CondDETR, xie2021segformer, liu2021swin}, we find that FLOPs are dominated by convolutions, not attention. % as in traditional transformers~\cite{devlin2018bert, dosovitskiy2020image, touvron2021training, chen2021crossvit}.
This result is due to two reasons: first, transformer models for detection often use CNN backbones to extract input features and second, many models have incorporated convolutions in the transformer encoder-decoder structure to achieve higher model accuracy and enable lower-cost attention. For complex vision applications beyond classification, it is crucial to analyze the substantial computation required for backbones and task-specific decoder heads.

Furthermore, we find that the distribution of FLOPs across model layers is not a good estimator of relative GPU execution time since GPU hardware and software have been well-optimized for exploiting the inherent locality and parallelism in convolutions~\cite{h100, tensorrt, chetlur2014cudnn}. This efficiency of convolutional operations is one reason for their continued use.  Figure~\ref{figure_detr_flops} shows these trends in FLOPs and GPU performance for detection models~\cite{carion2020end, meng2021-CondDETR, liu2022dabdetr, wang2022anchor} using a ResNet-50~\cite{he2016deep} backbone. % ResNet-50~\cite{he2016deep} as a CNN backbone. %% not attention as in early transformer models for natural language processing~\cite{vaswani2017attention, devlin2018bert} and image classification~\cite{dosovitskiy2020image, touvron2021training, chen2021crossvit} tasks.

Given the importance of efficiently executing attention \textit{and} convolutions for this class of models, we use MAGNet~\cite{venkatesan2019magnet}, an accelerator framework previously demonstrated for CNNs and attention-dominated transformers~\cite{keller202217}, to estimate the performance of these applications with customized hardware.

We then explore the impact of pruning in convolutional and attention layers in pretrained models on the model accuracy, execution time, and energy on an NVIDIA RTX A5000 GPU and the MAGNet-generated accelerator. Unintuitively, we find that larger models are not necessarily more resilient to accuracy loss when bypassing computation. Instead, how computation is split between the transformer encoder and decoder is a better indicator. 

Finally, we augment vision transformers with lower-cost execution paths that either leverage the resilience of pretrained models to pruning or rely on switching between retrained models for RDD inference. These paths achieve a range of execution time and energy savings, and can be alternatively executed to enable meeting dynamic resource constraints.

We make the following contributions:

\begin{itemize}
    
    \item We show that convolutions, not attention layers, dominate FLOPs for state-of-the-art vision transformer applications, since transformer models have integrated convolutions for accuracy and performance.

    \item We find that for these applications, the distribution of FLOPs across model layers is not a good estimator of relative GPU runtime. %Convolutions are well-suited for the parallel architecture in a GPU and have higher operational intensity compared to other model operations.

    \item We identify alternative lower-cost execution paths in these models and identify indicators for the resilience of pretrained vision transformers to dynamic pruning.
    
    \item We leverage a CNN accelerator framework and our dynamic computation bypassing approach to save 28\% of energy with a 1.4\% accuracy loss for SegFormer B2 with no additional training and 53\% of energy for RestNet-50 with a 3.3\% accuracy loss by switching between pretrained OFA models.
    
\end{itemize}
\section{Background}
\label{sec:background}

\subsection{RDD inference}

RDD inference targets a computational platform with a finite amount of resources~\cite{clemons2023leaf, wang2020dual}. Occasionally, there are not enough resources available for the full model execution, since there are other tasks that also need to be completed. In these scenarios, it is better to perform a degraded version of inference that requires less resources rather than to skip a frame and perform no inference. While statically selecting a model that matches the worst-case resource availability will ensure that no frames are missed, it does not allow for performing the full model execution and returning more accurate results when enough resources are available. Thus, the goal of RDD inference is to adjust the cost of inference to match the resources available for the computation, and thus maximize accuracy given input resource constraints.

\subsection{Vision Tasks and Models}

We consider semantic segmentation~\cite{yu2018methods} and object detection~\cite{zou2019object} since these tasks are widely used~\cite{alam2020survey, zhou2023survey, feng2020deep, siam2018comparative} and more complicated than image classification. Semantic segmentation assigns class labels to each pixel in an input image, while object detection identifies objects by providing bounding boxes and class labels.

% for computation section
\begin{table*}[t]
  {
  \centering

  \begin{tabular}{l|c|c|c|c|c|c} 

    Model & Task & Parameters (Millions) & Dataset & Input Image Size & GFLOPs & mIoU (SS) / AP (OD) \\
    
    \midrule
    
    SegFormer ADE B2~\cite{xie2021segformer} & SS & 28 & ADE20K~\cite{zhou2017scene} & 512 by 512 & 63 & 0.4651\\
    
    SegFormer City B2~\cite{xie2021segformer} & SS & 28 & Cityscapes~\cite{Cordts2016Cityscapes} & 1024 by 1024 & 290 & 0.8098\\
    
    Swin Tiny~\cite{liu2021swin} & SS & 60 & ADE20K & 512 by 512 & 237 & 0.4451\\
    
    Swin Small~\cite{liu2021swin} & SS & 81 & ADE20K & 512 by 512 & 259 & 0.4764\\
    
    Swin Base~\cite{liu2021swin} & SS & 121 & ADE20K & 512 by 512 & 297 & 0.4813\\
    
    \midrule
    
    DETR~\cite{carion2020end} & OD & 41 & COCO-2017~\cite{lin2014microsoft} & 800 by 1200 & 92 & 0.4200\\

    DAB DETR~\cite{liu2022dabdetr} & OD & 44 & COCO-2017 & 800 by 1200 & 97 & 0.328\\

    Anchor DETR~\cite{wang2022anchor} & OD & 37 & COCO-2017 & 800 by 1200 & 99 & 0.4188\\

    Conditional DETR~\cite{meng2021-CondDETR} & OD & 43 & COCO-2017 & 800 by 1200 & 96 & 0.4161\\
    
    % Deformable DETR~\cite{zhu2020deformable} & 40 & COCO~\cite{lin2014microsoft} & 800 by 1200 & 173 & 55.4 & 18 & 0.4454 & OD\\

  \end{tabular}

    \vspace{0.1in}

    \caption{Our state-of-the-art vision transformer case studies for semantic segmentation (SS) and object detection (OD). %Standard accuracy metrics are used: mIoU for SS and AP, with IoU from 0.5 to 0.95 in increments of 0.05, for OD.
    } 
    
    %SS = semantic segmentation. OD = object detection. % Standard accuracy metrics are used: mIoU for the ADE20K~\cite{zhou2017scene} and Cityscapes~\cite{Cordts2016Cityscapes} datasets (SS) and AP, with IoU from 0.5 to 0.95 in increments of 0.05, for the COCO-2017 dataset~\cite{lin2014microsoft} (OD).
    % \vspace{0.05in}

    \label{table_models}
  }

% \vspace{-0.3in}

\end{table*}

State-of-the-art models for these tasks have shifted to using transformer-based architectures, building on the success of using transformers for natural language processing~\cite{vaswani2017attention, devlin2018bert} and image classification~\cite{dosovitskiy2020image, touvron2021training, chen2021crossvit}. Most segmentation models and some detection models use the transformer encoder as a backbone for feature extraction, pairing it with different decoders and task-specific heads to produce the final output~\cite{wang2021pyramid, liu2021swin, zheng2021rethinking, zhang2022dino, zong2023detrs, zhou2022detecting}. Other detection models models employ a transformer encoder-decoder structure, often with a CNN backbone to extract visual features~\cite{carion2020end, liu2022dabdetr, meng2021-CondDETR, wang2022anchor, zhu2020deformable, zhang2022dino, zong2023detrs, zhou2022detecting}. Since we analyze the first type of models with our segmentation model case studies, we focus on the second type of models for our detection model case studies.

% % for computation section
% \input{tables/tbl_models}

% Since the latter is structurally and operationally similar to segmentation models we analyze, we focus on DETR-based models\cite{dai2021dynamic, zheng2020end, zhu2020deformable, yao2021efficient, roh2021sparse, liu2022dabdetr, wang2022anchor, meng2021-CondDETR, zhang2022dino, zong2023detrs, zhou2022detecting}.

% multi-scale feature fusion

Following from ViT, the first vision transformer, models reshape 2D images into a 1D sequence of flattened image patches, which are linearly projected into an embedding dimension~\cite{dosovitskiy2020image}. Throughout the model, layers progressively increase the embedding dimension and reduce the spatial dimensions to capture details at various image resolutions for higher accuracy. ViT is convolution-free since the original transformer layers consist of multi-head self-attention layers and multi-layer perceptrons (MLPs)~\cite{vaswani2017attention}. However, many modern vision transformers have incorporated convolutions with the transformer architecture for better accuracy results~\cite{xie2021segmenting, zheng2021rethinking, xie2021segformer, liu2021swin, carion2020end, liu2022dabdetr, meng2021-CondDETR, wang2022anchor, zhu2020deformable, kirillov2023segment}.

We use NVIDIA's SegFormer~\cite{xie2021segformer} and Microsoft's Swin Transformer~\cite{liu2021swin} as case studies for semantic segmentation since they achieve state-of-the-art accuracy and are commonly used. Figure~\ref{figure_segformer_swin_models} shows these model architectures. These models build from the encoder in ViT, and further optimize attention and integrate convolutions to preserve local continuity information. Both models have four encoder stages followed by a decoder. SegFormer B2 has three, four, six, and three encoder blocks in stages zero through three, respectively. Swin Tiny has six encoder blocks in stage two and two encoder blocks in all the other encoder stages. Individual layers inside these blocks are shown in Figure~\ref{figure_segformer_swin_models}, with breakout boxes showing the individual layers in the MLP and attention components.

Prior work has focused on improving encoder backbones~\cite{wang2021pyramid, liu2021swin} that can be used with different decoders and task-specific heads for various vision tasks. In these backbones, attention dominates FLOPs~\cite{liu2021swin, xie2021segformer, dosovitskiy2020image, touvron2021training, chen2021crossvit}, so prior work has heavily focused on reducing the cost of attention~\cite{li2022divit, wang2022towards, you2022vitcod, xiao2018unified}. SegFormer contributes a simpler attention-free decoder, while Swin directly uses the UPerNet decoder head~\cite{xiao2018unified}. Both of these decoders use convolutions to extract and fuse features from different spatial resolutions and reduce channel dimensions, which results in higher accuracy.

For object detection, many models build from the DETR architecture~\cite{carion2020end}. DETR uses a CNN backbone to extract image features, a conventional transformer, and a feed forward network to return the detection prediction. We examine DETR and three recent models based on DETR: DAB DETR~\cite{liu2022dabdetr}, Anchor DETR~\cite{wang2022anchor}, and Conditional DETR~\cite{meng2021-CondDETR}. These models improve aspects of the transformer in DETR. We consider the base variants of these models, as included in detrex, an open-source toolbox for detection models~\cite{ideacvr2022detrex}. The base versions of these models use ResNet-50~\cite{he2016deep} as the CNN backbone.

% For object detection, transformer models usually come in two forms. One type builds from the DETR architecture, which uses a CNN backbone to extract image features, a conventional transformer, and a FNN to return the detection prediction~\cite{carion2020end}. %Note that many non-transformer models for detection also employ a CNN backbone. 
% The other type uses a transformer encoder backbone, such as Swin~\cite{liu2021swin}, and a task-specific head with multi-scale feature fusion. Since the latter is structurally and operationally similar to segmentation models we analyze, we focus on DETR-based models\cite{dai2021dynamic, zheng2020end, zhu2020deformable, yao2021efficient, roh2021sparse, liu2022dabdetr, wang2022anchor, meng2021-CondDETR, zhang2022dino, zong2023detrs, zhou2022detecting}. We examine DETR and three recent models based on DETR: DAB DETR~\cite{liu2022dabdetr}, Anchor DETR~\cite{wang2022anchor}, and Conditional DETR~\cite{meng2021-CondDETR}. These models improve aspects of the transformer in DETR. We consider the base variants of these models, as included in detrex, an open-source toolbox for detection models~\cite{ideacvr2022detrex}. The base versions use ResNet-50~\cite{he2016deep} as the CNN backbone.

Prior work has also explored the resiliency of ResNet-50~\cite{cai2019once, jastrzkebski2017residual}. One example is OFA~\cite{cai2019once}, which trains the model once to produce many sets of weights for different model subsets, resulting in models that significantly reduce FLOPs with modest decreases in accuracy. In Section~\ref{sec:resiliency_ofa}, we leverage switching between OFA ResNet-50 models to enable RDD inference. We denote the most accurate OFA ResNet-50 model as OFA-ResNet-50.

Finally, we use standard accuracy metrics in our analysis: mean intersection over union (mIoU) for semantic segmentation and average precision (AP) for object detection. IoU is the area of overlap between the predicted segmentation and the ground truth divided by their total area. Thus, mIoU is the average of the IoU for every class. AP measures the precision-recall curve across different IoU confidence thresholds, which specify the minimum IoU to consider a positive match. For the COCO dataset~\cite{lin2014microsoft}, AP is the average precision for IoU from 0.5 to 0.95, in increments of 0.05. Both mIoU and AP range from zero to one, and higher values indicate better accuracy.
\section{Computation in Vision Transformers}
\label{sec:vt_computation}

Table~\ref{table_models} provides an overview of our model case studies. To understand the computation required for inference with modern vision transformer applications, we profile the FLOPs and GPU execution time distribution among model layers. We use an NVIDIA RTX A5000 GPU. 

Unlike prior work that focuses on classification and a 224 by 224 image size (49K pixels)~\cite{imagenet_cvpr09}, our evaluation is across various image sizes (1K to 4M pixels), including those used in standard datasets for classification (ImageNet~\cite{imagenet_cvpr09}), segmentation (ADE20K~\cite{zhou2017scene}, Cityscapes~\cite{Cordts2016Cityscapes}), and detection (COCO~\cite{lin2014microsoft}, Cityscapes~\cite{Cordts2016Cityscapes}). These results become important for emerging applications that call for larger image resolutions~\cite{yu2018methods, zou2019object}.

\subsection{FLOPs in Semantic Segmentation Models}
\label{sec:computation_ss}

\begin{figure}[t]

  \centering

  \includegraphics[width=1\linewidth]{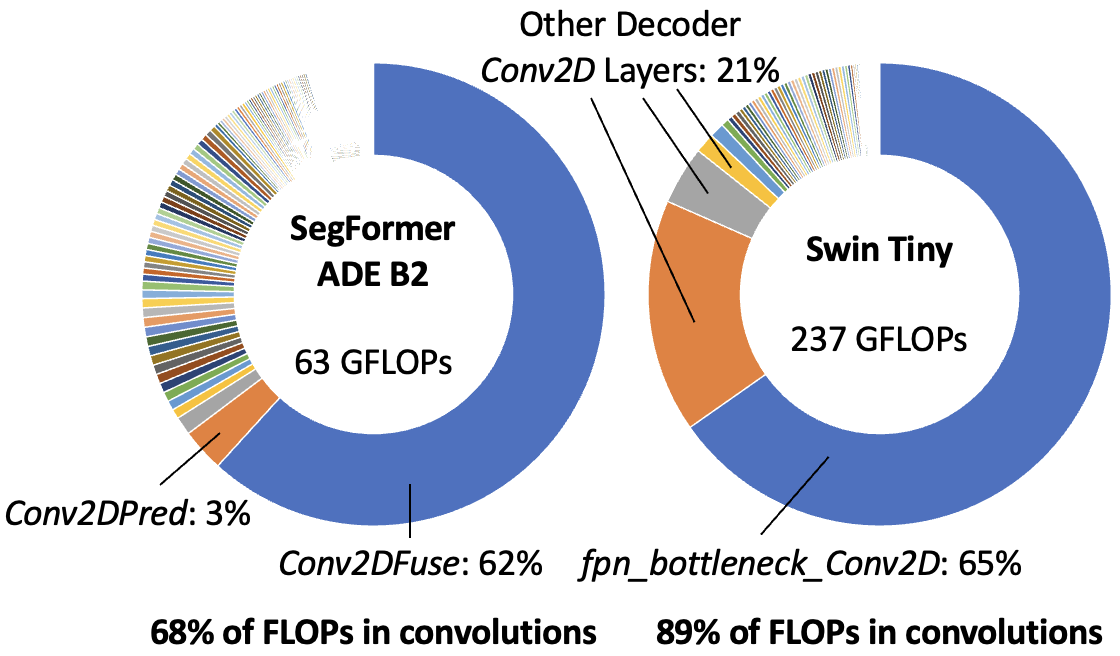}
  
  % \vspace{-0.05in}
  
  \caption{FLOPs distribution across SegFormer ADE B2 model layers and Swin Tiny model layers for inference with a 512 by 512 input image size. %Convolutions are 68\% and 89\% of FLOPs.
  }
  
  \label{figure_segformer_b2_computation}
  
  % \vspace{-0.15in}
 
\end{figure}

% To quantify model accuracy, we use a common accuracy metric for semantic segmentation: mean intersection over union (mIoU). IoU is the area of overlap between the predicted segmentation and the ground truth divided by their total area. Thus, mIoU is the average of the IoU for every class and ranges from zero to one. Higher values indicate more overlapping segmentation and thus better accuracy.

We show the distribution of FLOPs across model layers for SegFormer B2 and Swin-Tiny with a 512 by 512 image size in Figure~\ref{figure_segformer_b2_computation}. We find that 68\% and 89\% of the total FLOPs are in convolution layers in SegFormer B2 and Swin-Tiny, respectively, in stark contrast to the zero convolutions in ViT~\cite{dosovitskiy2020image} and BERT~\cite{devlin2018bert}. Relatively large convolutional layers are labeled in the FLOPs distributions in Figure~\ref{figure_segformer_b2_computation}. While encoders use convolutions to overlap image patches to incorporate local continuity information and reduce the size of inputs for attention, only 5\% of convolution FLOPs are in the SegFormer encoder and 1\% are in Swin encoder. Instead, as shown in Figure~\ref{figure_segformer_swin_models}, decoders replace attention with large convolutions to fuse information at different image resolutions from the output of each encoder stage. We denote these convolutions as \textit{Conv2DFuse} in SegFormer and \textit{fpn\_bottleneck\_conv} in Swin.

% This layer's function in this model is to fuse information about the input image at various image resolutions from the output of each encoder stage.

% In SegFormer, 67\% of the FLOPs are in the decoder and 95\% of convolutions are in the decoder, while in Swin, 89\% of the FLOPs are in the decoder and 99\% of the convolutions are in the decoder. Convolutions in the encoder are used to overlap image patches to incorporate local continuity information and reduce the size of inputs for attention to reduce the complexity of this layer. In the decoder, 

Looking at SegFormer's decoder, where nearly 70\% of the FLOPs are used, the \textit{Conv2DFuse}, \textit{Conv2DPred}, and \textit{Decoder Linear connected to output of encoder stage zero} (denoted as \textit{DecodeLinear0}) layers individually comprise $62\%$, $3\%$, and $1.3\%$ of the total number of FLOPs for a 512 by 512 image, respectively, as shown in Figure~\ref{figure_segformer_b2_computation}. In particular, \textit{Conv2D Fuse} alone constitutes the majority of FLOPs, considerably more so than any other layer in the model. The \textit{Conv2D Fuse} layer has 3072 input channels, 768 output channels, and a 1 by 1 stencil.

When considering the Swin Tiny model for segmentation, we find that 89\% of FLOPs are in the decoder, and 99\% of convolutions are in this decoder. The \textit{fpn\_bottleneck\_Conv2D} layer alone comprises 65\% of FLOPs. This layer has 2048 input channels, 512 output channels, and a 3 by 3 stencil. 

From these results, it follows that these segmentation models have high operational intensity, at 130+ operations/byte, and that they are not memory bound like traditional transformers for language~\cite{devlin2018bert} and classification~\cite{dosovitskiy2020image, touvron2021training} tasks. We also see that the task-specific decoder heads are computationally dominant for transformer models for segmentation, and critical to include in profiling analyses for this task.

\begin{figure}[t]

  \centering

  \includegraphics[width=1\linewidth]{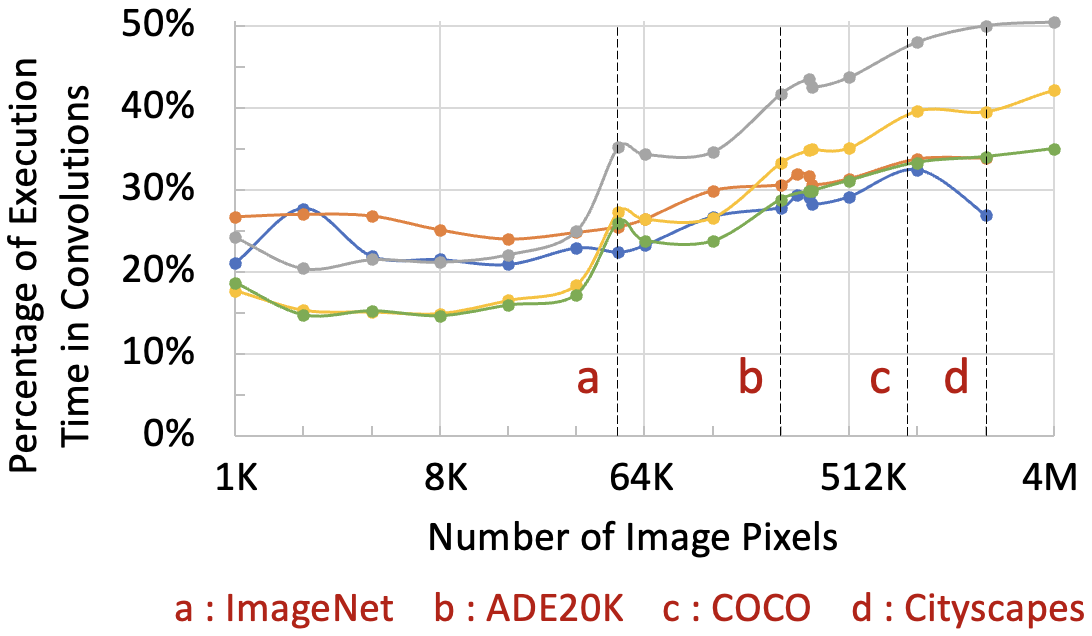}

  % \vspace{-0.1in}
  
  \caption{Image pixels versus NVIDIA RTX A5000 GPU execution time spent on convolutions for inference with the SegFormer ADE B2 (blue), SegFormer City B2 (orange), Swin Tiny (gray), Swin Small (yellow), and Swin Base models (green).
  }
  
  \label{figure_segformer_b2_image_sizes}
  
  % \vspace{-0.15in}
 
\end{figure}

\subsection{FLOPs in Object Detection Models}
\label{sec:computation_detection}

Figure~\ref{figure_detr_flops} shows the FLOPs distribution between convolution and non-convolution layers as well as the FLOPs distribution between the ResNet-50 backbone and transformer for DETR-based models across various image sizes. Note that the percentage of FLOPs in convolutions and in ResNet-50 are very similar, indicating that the convolutions in DETR-based models are predominantly in the ResNet-50 backbone (and that the contribution of non-convolutional layers to the ResNet-50 backbone FLOPs is negligible as expected). Solely focusing on smaller image sizes below 64K pixels would lead one to conclude that improving the transformer is important for overall model efficiency. However, for object detection, we typically need to consider much larger image sizes. The importance of the ResNet-50 backbone compared to the transformer mostly increases with larger image sizes: for more than 128K image pixels, the ResNet-50 backbone comprises at least half of the total FLOPs in all of these models, and for more than 1M pixels, which corresponds to standard image sizes for detection datasets~\cite{lin2014microsoft, Cordts2016Cityscapes}, this backbone requires 80+\% of FLOPs.

% For the average image size in the commonly-used COCO dataset~\cite{lin2014microsoft} for detection, all these DETR-based models have similar results: the ResNet-50 backbone is instead computationally dominant, requiring 80 to 86\% of the total FLOPs. Furthermore, t

% \subsection{Key Application Insights}

% Modern vision transformer models majorly comprise of convolutions, not attention, due to efforts to build simpler decoders and the use of CNN backbones, in contrast to earlier attention-dominated transformer models such as BERT~\cite{devlin2018bert} and ViT~\cite{dosovitskiy2020image}. Furthermore, the proportion of FLOPs and GPU runtime in convolutional layers increases for the larger image sizes required for more complicated vision tasks like segmentation and detection. Thus, models overall have relatively high operational intensity, at 130+ operations/byte, and are not memory-bound like traditional transformers.

\subsection{Using FLOPs to Estimate GPU Runtime}

We observe that the number of FLOPs in all of these models does not directly map to GPU execution time, as shown in Figure~\ref{figure_segformer_b2_image_sizes} for the segmentation models and Figure~\ref{figure_detr_flops} for the detection models. For 512 by 512 image sizes, convolutions comprise 28\% and 42\% of the total execution time for SegFormer B2 and Swin Tiny, respectively, despite requiring 68\% and 89\% of the FLOPs. With the DETR-based models, 30\% to 40\% of the total execution time for larger image sizes is in convolutions, which make up 80+\% of the total FLOPs. For all models, the importance of convolutions generally increases with larger image sizes, but the proportion of GPU runtime spent in convolutions is much lower than the relative FLOP counts. These results show that the GPU implementation is well-designed for reusing convolution weights and exploiting convolution parallelism. Thus, integrating convolutions results in better model accuracy \textit{and} performance. 

Figure~\ref{figure_segformer_b2_image_sizes} shows that convolutions comprise a smaller portion of FLOPs and GPU performance in larger Swin models compared to Swin Tiny. This result makes sense because all models have a similar decoder, which majorly comprises of convolutions, but Swin Small and Swin Base have two times the number of encoder blocks compared to Swin Tiny. We also observe that matrix multiplications comprise about an equal proportion of execution time compared to convolutions on the GPU for SegFormer and Swin for larger image sizes commonly used for segmentation. %, since both these operations can be easily parallelized with the parallel hardware in a GPU.

Thus, while many prior papers use FLOPs to estimate the runtime savings for inference~\cite{wang2018skipnet, xin2020deebert, kaya2019shallow, hu2020triple, graves2016adaptive, liu2020fastbert, figurnov2017spatially}, we find that FLOPs alone is not a good predictor of runtime on a GPU. GPU runtime is instead a function of FLOPs, layer parallelism, and layer operational intensity. Convolutions enable spatial parallelism since each output pixel depends on a relatively small receptive field in the input image. As a result, these layers can be efficiently executed on the highly parallel architecture in a GPU. Attention can be computed in parallel for several image patches, enabling temporal parallelism, and matrix multiplication can be easily parallelized. However, attention mechanisms also involve sequential operations where these patches attend to each other, making the overall operation less parallelizable. In addition, attention layers in these vision transformers have lower operational intensity compared to convolutions, resulting in more memory accesses. Thus, it is important for inference engines for vision transformers to efficiently execute CNNs and attention.

\section{Profiling on a Hardware Accelerator}
\label{sec:accelerator}

We next analyze the performance and energy of these applications on the type of compute that one is likely to see in embedded platforms and real-time systems for RDD inference. 
For applications, we use SegFormer B2 and Swin Tiny for segmentation with a 512 by 512 input image size. Given the reliance of DETR-based models on the ResNet-50 backbone from Section~\ref{sec:vt_computation}, we include OFA-ResNet-50 with a 224 by 224 input image size. These applications have a range of model sizes (28 to 60 million parameters), computation requirements (4 to 237 GLOPs) and model structures (68\% to 95+\% of FLOPs in convolutions). 
%
% With these applications, we explore throughput, area, and energy tradeoffs between accelerators with different compute organization and memory sizes. 
With these applications, we show how to tune CNN accelerators for modern vision transformers by exploring throughput, area, and energy tradeoffs with different compute organization and memory sizes. This profiling lays the groundwork for executing dynamic vision transformers on hardware accelerators in Section~\ref{sec:resiliency}.

\subsection{Hardware Overview}
\label{sec:hw_overview}

\begin{figure}[t]

  \centering
  
  \includegraphics[width=1\linewidth]{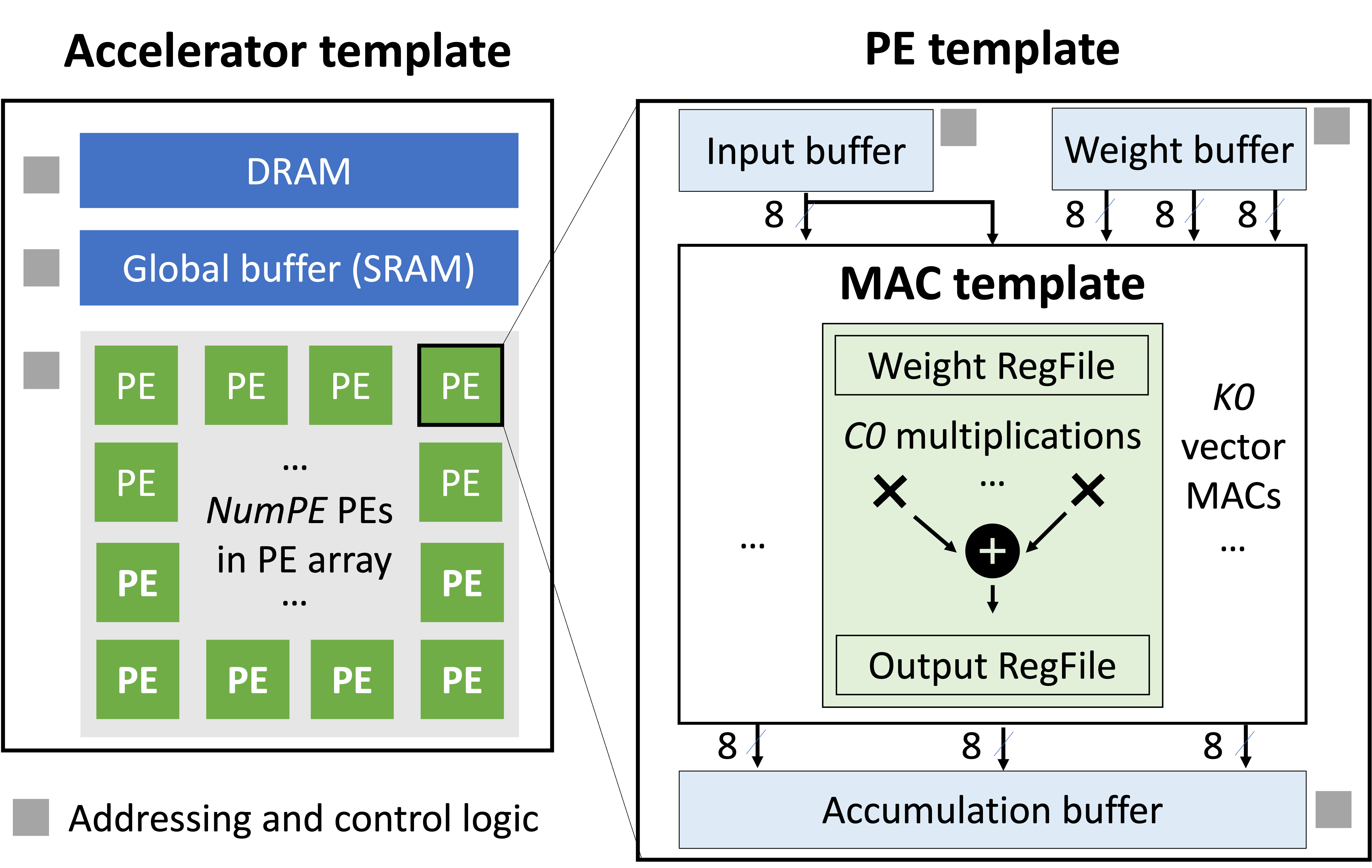}

  % \vspace{-0.05in}
  
  \caption{Parameterizable MAGNet~\cite{venkatesan2019magnet} accelerator and PE architecture templates.}
  
  \label{figure_magnet}
  
  \vspace{-0.15in}
 
\end{figure}

We use MAGNet~\cite{venkatesan2019magnet}, a CNN accelerator framework, which has been previously extended to execute attention-dominated transformers for language and classification tasks as well~\cite{keller202217}. As a result, its parameterizable accelerator template provides a reasonable architecture for profiling our applications on a hardware accelerator. Most other accelerators for transformers focus on attention and custom detection components~\cite{li2022divit, wang2022towards, you2022vitcod, zeng2022fpga}, often without support for convolutions, and are thus not well-optimized for more recent vision transformers.

Figure~\ref{figure_magnet} shows a block diagram of the MAGNet accelerator and PE architecture templates. The accelerator has three levels of compute with vector multiply-accumulate (MAC) units, processing elements (PEs), and the PE array. All levels allow for parallel execution in order to exploit data parallelism present in model layers: each vector MAC unit can execute $C0$ multiplications in parallel, every PE has $K0$ parallel vector MAC units, and the PE array has $NumPE$ PEs. Each PE also contains a post-processing unit to fuse activation and pooling layers with the preceding convolutions.

The memory hierarchy has four levels: small register files within each vector MAC unit (labeled as ``RegFile" in Figure~\ref{figure_magnet}), local buffers within each PE, a global buffer at the PE array level, and an off-chip DRAM. Inputs are shared across the vector MAC units with one input buffer per PE. Each vector MAC unit has its own weight buffer, with $K0$ total weight buffers in every PE. We can temporally tile the weights (split by output channels) and activations (split by image height and width). The paper on the MAGNet framework provides further detail on how convolutions and matrix multiplications can be mapped to this accelerator template~\cite{venkatesan2019magnet}.

MAGNet uses Mentor Graphics Catapult HLS to generate RTL from a synthesizable SystemC and C++ architecture description. We report results in a 5nm technology, using Synopsys Design Compiler for synthesis for execution time and area results and Primetime-PX with Gaussian synthetic data for power results.

Since prior work has shown that dataflow choice minimally affects energy efficiency~\cite{yang2020interstellar}, we directly use an output-stationary local-weight-stationary dataflow that was previously shown to work well for CNNs~\cite{venkatesan2019magnet} and attention layers~\cite{keller202217}. Like these prior works, we also use 8-bit precision for data.

\begin{table}[t]
  {
  \centering
  \setlength{\tabcolsep}{3pt}

  \begin{tabular}{c|c|c|c|c|c} 

    Label & NumPE & K0 = C0 & Weight buffer & Input buffer & PE array\\
    & & & size (kB) & size (kB) & area (mm$^2$)\\
    
    \midrule

    A & 32 & 32 & 1024 & 64 & 16.7\\
    B & 32 & 32 & 128 & 64 & 4.5 \\

    \midrule
    
    C & 16 & 32 & 1024 & 64 & 8.3 \\
    D & 16 & 32 & 128 & 64 & 2.3 \\
    E & 16 & 32 & 128 & 32 & 1.9 \\
    F & 16 & 32 & 64 & 64 & 2.0 \\
    G & 16 & 32 & 64 & 32 & 1.7 \\

    \midrule

    H & 64 & 16 & 128 & 32 & 6.1 \\
    I & 64 & 16 & 128 & 16 & 5.4 \\
    J & 64 & 16 & 64 & 32 & 4.2 \\
    K & 64 & 16 & 64 & 16 & 3.5 \\
    L & 64 & 16 & 32 & 32 & 3.3 \\
    M & 64 & 16 & 32 & 16 & 2.6 \\
    
  \end{tabular}
  
    \vspace{0.1in}
    
    \caption{The MAGNet accelerator parameterizations we explore.}
    
    % Template parameter values and PE array area for accelerators explored.
    
    % Accelerators explored, their corresponding template parameter values, and PE array areas.
    % \vspace{0.05in}
    \label{table_accelerator_configs}
  }

% \vspace{-0.3in}

\end{table}

\begin{figure}[t]

  \centering

  \includegraphics[width=1\linewidth]{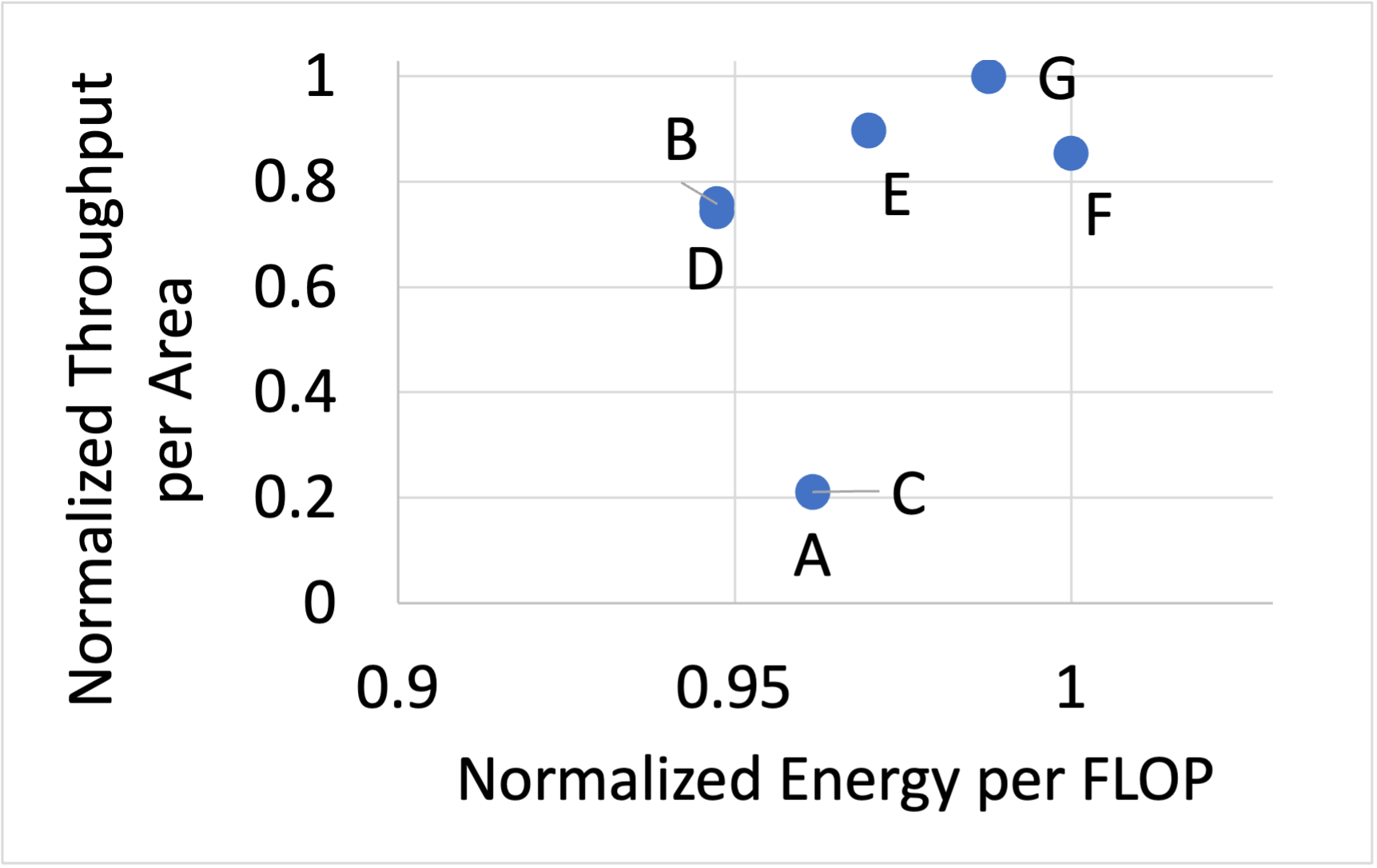}

  % \vspace{-0.05in}
  
  \caption{
  Energy per FLOP versus throughput per area for SegFormer ADE B2 inference with a 512 by 512 input image size. Labels indicate the hardware accelerators in Table~\ref{table_accelerator_configs}. 
  % with $K0 = C0 = 32$.
  }
  
  \label{figure_segformer_throughput_energy_area}
  
  % \vspace{-0.175in}
 
\end{figure}

\subsection{Tradeoffs with Different Accelerator Parameters}

To examine throughput, area, and energy tradeoffs for vision transformer applications, we compare different accelerator template parameters, as shown in Table~\ref{table_accelerator_configs}. Accelerators C through M can compute the same number of MACs in parallel, while accelerators A and B have twice the compute capability. To fairly compare the execution of this class of models on these parallel, throughput-focused compute engines, we compute the energy per operation, which to first order is independent of parallelism, and throughput/mm$^2$, to normalize throughput by the silicon cost of its hardware. As a result, accelerator B has twice the available compute compared to accelerator D and is twice as fast, but it looks identical to accelerator D with our metrics since it is also twice as big.

Figure~\ref{figure_segformer_throughput_energy_area} explores the tradeoff between these metrics for SegFormer ADE B2, where lower energy per operation and higher throughput per area signify more efficient inference. We also checked Swin Tiny and OFA-ResNet-50, which have similar results compared to SegFormer. All of these models can significantly reuse data across the channel dimension. As a result, accelerators with reduced vectorization ($K0 = C0 = 16$) result in lower efficiency (1.4$\times$ more energy per FLOP and 2.8$\times$ more area per FLOP) compared to those with $K0 = C0 = 32$. Since the accelerators with $K0 = C0 = 16$ are not competitive with either metric, they are not shown in Figure~\ref{figure_segformer_throughput_energy_area}.

% Accelerators with reduced vectorization ($K0 = C0 = 16$) are not shown in Figure~\ref{figure_segformer_throughput_energy_area} since they are not competitive with either of these metrics for any of the models: there is less opportunity to reuse partial sums spatially (1.4$\times$ more energy per FLOP) and any throughput gains are lost from the large increase the area (2.8$\times$ more area per FLOP). 

Swin Tiny has significantly more model layers with an odd number of input and/or output channels (such as 49), due to default layer parameters or how matrix multiplications are mapped to the hardware. These channel counts are not divisible by $K0$ and $C0$ whether they are equal to 16 or 32, resulting in similar underutilization in the PEs for all the accelerator parameterizations, and similar performance across accelerators. In contrast, SegFormer and OFA-ResNet-50 have more evenly divisible channel values, resulting in slightly higher utilization and 10\% faster execution with accelerators with $K0 = C0 = 16$ compared to those with $K0 = C0 = 32$.

For all of these models, accelerators D, E, and G are Pareto-optimal with our metrics. We expect this trend to continue for larger models, including CNN backbones such as ResNet-101 for DETR-based models. Note that accelerators E and G are the two smallest accelerators in Table~\ref{table_accelerator_configs} as well. Not surprisingly, these accelerators are similar to accelerator D, which was the design previously used for transformer models dominated by attention~\cite{keller202217}. 

% for performance distributions subsection
\begin{figure}[t]

  \centering
  
  \includegraphics[width=1\linewidth]{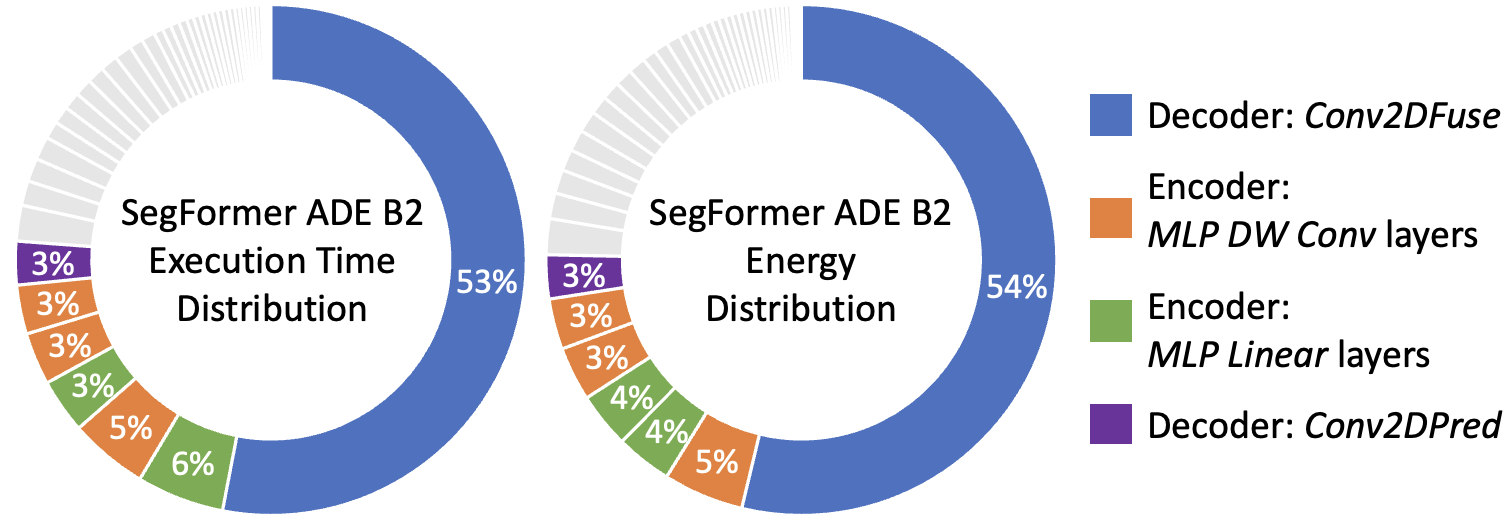}

  % \vspace{-0.1in}
  
  \caption{Execution time and energy distributions across layers in the SegFormer ADE B2 model for inference with a 512 by 512 input image size on accelerator E. %Layers comprising at least 3\% of time/energy are labeled.
  }
  
  \label{figure_segformer_full_time_energy}
  
  % \vspace{-0.1in}
 
\end{figure}

\begin{figure}[t]

  \centering
  
  \includegraphics[width=1\linewidth]{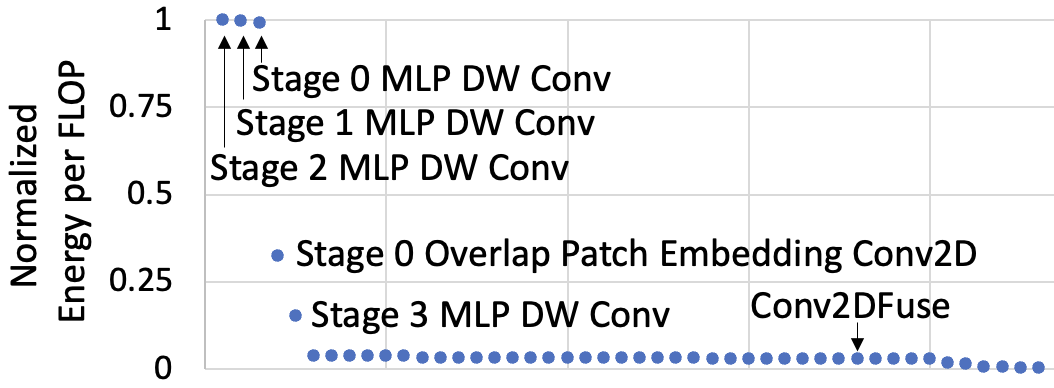}

  % \vspace{-0.05in}
  
  \caption{Normalized energy per FLOP for SegFormer ADE B2 model layers for inference with a 512 by 512 input image size on accelerator E. The layer with highest energy per FLOP is set to one, with all layers sorted in order of decreasing energy per FLOP from left to right.
  }

\label{figure_segformer_energy_per_flop}
  
  % \vspace{-0.2in}
 
\end{figure}

In Section~\ref{sec:vt_computation}, we saw that these vision transformers have higher operational intensity than attention-dominated transformers, due to the increased locality of the convolution operation. Thus, it makes sense that the smaller memories in accelerators E and G result in higher throughput per area compared to accelerator D. We find that 64 to 128 B/MAC and 32 to 64 B/MAC are sweet spots for the weight and input buffers, respectively, to balance throughput and area tradeoffs for these applications.

\subsection{Performance Distributions}
\label{sec:hw_eval}

To balance energy and area constraints, we profile our model case studies on accelerator E, which is between accelerators D and G on the Pareto curve in Figure~\ref{figure_segformer_throughput_energy_area}. The synthesized clock frequency for accelerator E is 1.25 GHz.

The SegFormer ADE B2 model runs in 3.6ms on accelerator E. %, which is 6.7$\times$ faster than on the NVIDIA RTX A5000 GPU.
Figure~\ref{figure_segformer_full_time_energy} shows the execution time and energy breakdowns across model layers for inference on accelerator E. Convolutions constitute 74\% of the total execution time and energy. The largest layer, \textit{Conv2DFuse}, alone requires over half of the total execution time and energy on accelerator E. % This proportion with specialized hardware better reflects the 68\% of model FLOPs in convolutional layers compared to the 30\% of GPU execution time spent in convolutions.

\begin{figure}[t]

  \centering
  
  \includegraphics[width=1\linewidth]{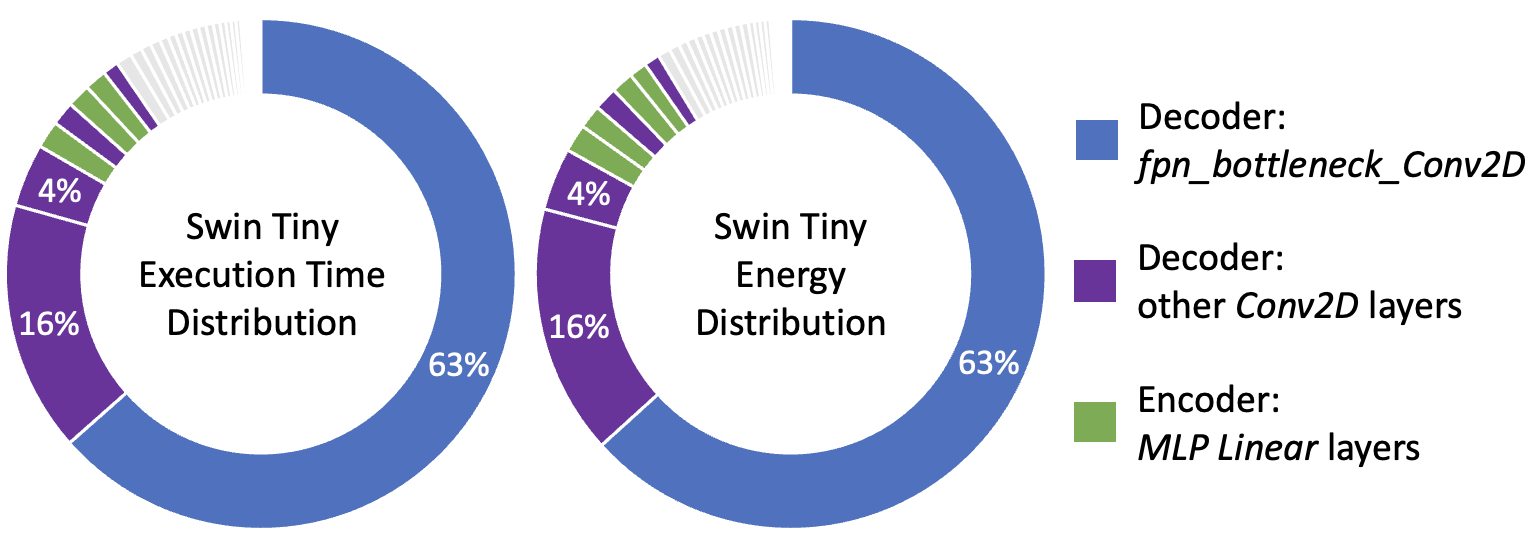}

  % \vspace{-0.1in}
  
  \caption{Execution time and energy distributions across layers in the Swin Tiny model for inference with a 512 by 512 input image size on accelerator E. %Layers comprising at least 3\% of time/energy are labeled.
  }
  
  \label{figure_swin_full_time_energy}
  
  % \vspace{-0.2in}
 
\end{figure}

Five convolution layers in the encoder have higher energy per FLOP compared to other layers, as indicated in Figure~\ref{figure_segformer_energy_per_flop}. Fortunately, these layers comprise only 14\% of the total energy. These layers have a small number of input channels: the \textit{Overlap Patch Embedding Conv2D} layer has three input channels for an input image to the model, and the \textit{MLP DW Conv} layers each have one input channel due to how we exploit parallelism in mappings for depthwise convolutions. This results in hardware underutilization, since all 32 parallel multiplications in a vector MAC unit are not utilized as often as in other layers. Thus, convolutions comprise a slightly greater portion of accelerator time and energy (74\%) compared to model FLOPs (68\%). Note that \textit{Conv2DFuse}, with 3072 input channels, more fully utilizes available parallelism in the vector MAC units and thus has relatively low energy per FLOP.

The Swin Tiny model runs in 12ms on accelerator E. %, which is 4.5$\times$ faster than on the NVIDIA RTX A5000 GPU.
The Swin Tiny execution time and energy distributions on accelerator E, shown in Figure~\ref{figure_swin_full_time_energy}, closely match the model FLOPs distribution: 87\% of the execution time and energy on E are in convolutions, compared to the 89\% of FLOPs in convolutions. The distribution of time and energy on accelerator E across individual layers also match the distribution of FLOPs across model layers. For example, the largest layer, \textit{fpn\_bottleneck\_conv}, comprises 63\% of the total execution time and energy on accelerator E, and 65\% of model FLOPs.

Swin Tiny layers with relatively higher energy per FLOP are either convolutions in the decoder with a low number of input channels, due to how we map this computation to maximize the available parallelism, or matrix multiplications in attention blocks, due to odd numbers of input channels that do not fully utilize available parallelism, leading to lower utilization.

With OFA-ResNet-50, the execution time and energy distributions for inference on accelerator E are mostly evenly split among all the convolutions in the CNN. Two layers have significantly higher energy per FLOP compared to the remaining layers. These layers are the first layer (the input layer) and the last layer (the classifier) in the model. The first layer has three input channels for an input image to the model, while the last layer has one input channel for a linear layer. Thus, as with the other models, these layers with a small number of input channels have lower utilization in the vector MAC units.

\section{Enabling RDD Inference}
\label{sec:resiliency}

% To enable RDD inference with these models, we use our profiling results on the A5000 GPU in Section~\ref{sec:vt_computation} and on accelerator E from Section~\ref{sec:accelerator} to explore the resulting performance and energy on these platforms when modulating the computation in the most critical model layers.
To enable RDD inference with these models, we explore modulating the computation in the most critical layers in order to examine the resulting performance and energy on the A5000 GPU and on accelerator E from Section~\ref{sec:accelerator}.

For the segmentation models, we examine model resilience to selectively bypassing computation in encoder blocks and convolutional layers with pretrained model weights as well as the execution time and energy savings achievable by switching between retrained model weights. For the pretrained model experiments, we various sweep model and layer parameters and run inference with the corresponding subset of pretrained model weights from the original model. We analyze SegFormer in Section~\ref{sec:resiliency_seg} and Swin Transformer in Section~\ref{sec:resiliency_swin}. 
For the DETR-based models, we leverage switching between OFA ResNet-50 models in Section~\ref{sec:resiliency_ofa} to modulate the computation in the CNN backbone. 

Based on these explorations, we identify general principles to guide the search for identifying competitive alternative execution paths for vision transformer models in Section~\ref{sec:resiliency_principles}.
Finally, Section~\ref{sec:resiliency_rdd} briefly overviews how we can use these results for RDD inference.

\subsection{SegFormer B2 Model Resilience and Switching}
\label{sec:resiliency_seg}

We examine the resilience of SegFormer B2 with different image sizes, using models trained and validated with 512 by 512 images in the ADE20K dataset~\cite{zhou2017scene} (denoted ``SegFormer ADE B2") and with 1024 by 1024 images in the Cityscapes dataset~\cite{Cordts2016Cityscapes} (denoted ``SegFormer City B2"). Since the majority of FLOPs are in the decoder for both models, we find that the model accuracy is not resilient to reducing scaling factors and input channels in individual layers in the encoder, making these modified execution paths ill-suited for RDD inference.

Next, we examine the impact of reducing the number of input and output channels for the top three layers in the FLOPs distribution in Figure~\ref{figure_segformer_b2_computation}: \textit{Conv2DFuse}, 
\textit{Conv2DPred}, and \textit{DecodeLinear0}. Reducing the number of input channels to \textit{DecodeLinear0} does not enable skipping computation in earlier layers since the full output of encoder stage zero must still be computed as input for encoder stage one. Similarly, since the interpolation of the output of the \textit{Linear} layers after the encoder stages is reversed and then concatenated, reducing input channels from 3072 to \textit{Conv2DFuse} only allows for skipped computation to propagate backwards if the model execution path uses less than or equal to 768 input channels (corresponding to the contribution from encoder stage three, which is not input to another encoder stage and only input to \textit{Conv2DFuse}). Since the decoder layer outputs have only one following destination, skipped computation associated with input channels to \textit{Conv2DPred} can be propagated backwards through preceding decoder layers (such as \textit{ReLU}) to further reduce FLOPs.

% for SegFormer B2 resilience subsection
\begin{figure}[t]

  \centering
  
  \includegraphics[width=1\linewidth]{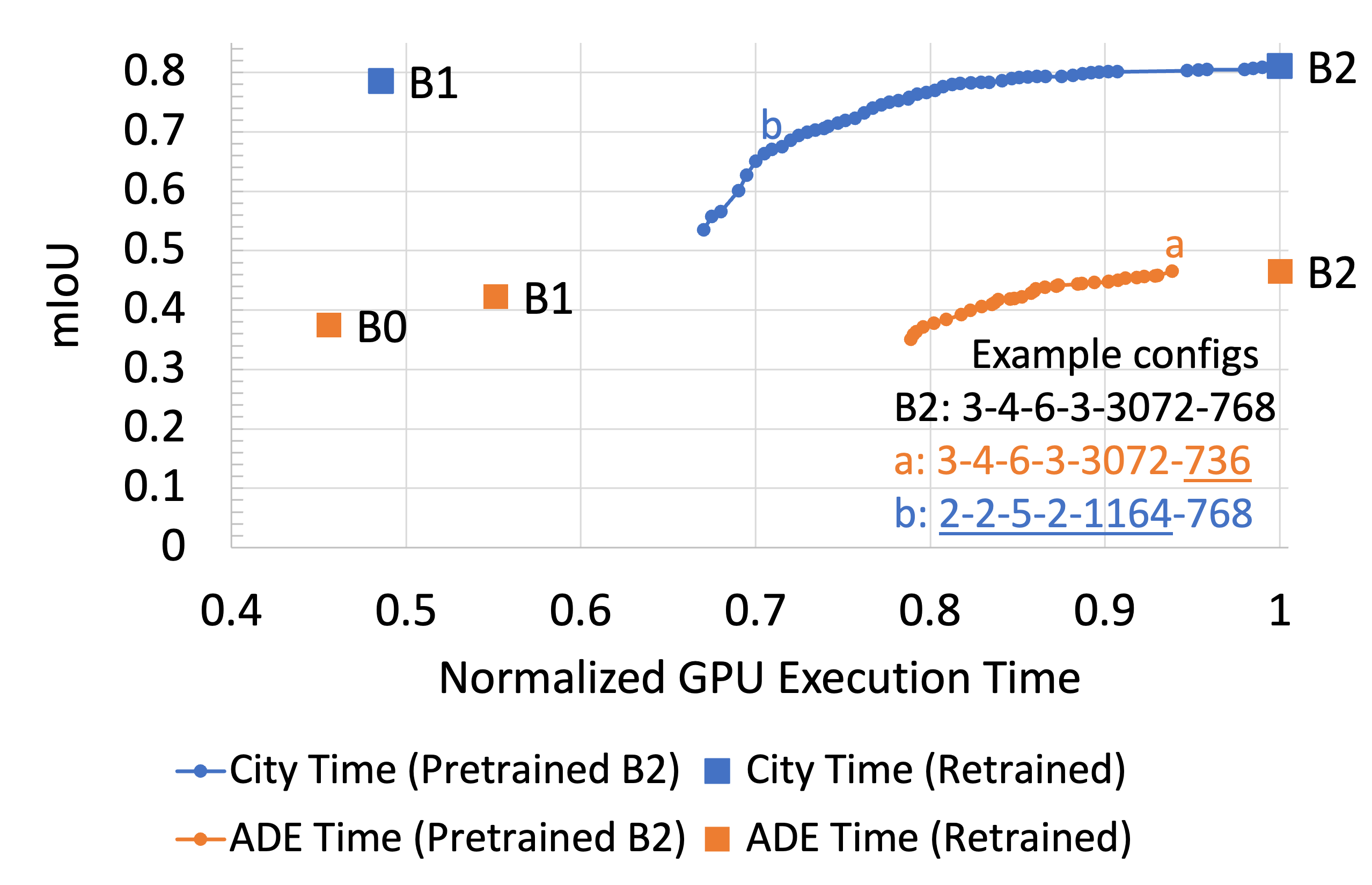}

    % \vspace{-0.1in}
  
  \caption{Tradeoff between execution time and mIoU (model accuracy) when dynamically pruning pretrained SegFormer B2 models on the NVIDIA RTX A5000 GPU (original retrained model executions on the GPU shown as squares). Example dynamic configuration labels list the number of encoder blocks in each of the four encoder stages followed by the input channels for \textit{Conv2DFuse} and \textit{Conv2DPred}; values changed from the original model are underlined.
  }
  
  \label{figure_segformer_pareto_gpu}
  
  % \vspace{-0.15in}
 
\end{figure}

Given a constant number of bypassed channels $x$, we found that which channels that were removed (e.g., the first $x$ versus the last $x$ versus the smallest $x$) did not substantially change the resulting accuracy, so we bypass the last $x$ in our experiments. As more channels are removed from the inputs to \textit{Conv2DFuse}, \textit{Conv2DPred}, and \textit{DecodeLinear0}, the marginal accuracy drop increases. However, if we additionally bypass full encoder blocks, we can shift these trade-off curves more to the left than down (saving relatively more execution time compared to the relative accuracy drop), and produce new Pareto-optimal execution paths. Thus, Figure~\ref{figure_segformer_pareto_gpu} shows the piecewise concatenation of the Pareto-optimal points on these shifted curves for SegFormer ADE B2 and SegFormer City B2 on the A5000 GPU. 

We find that combinations of varying the number of encoder blocks in each stage and input channels to \textit{Conv2DFuse} and \textit{Conv2DPred} result in Pareto-optimal points, as shown with the ``ADE Time (Pretrained B2)" and ``City Time (Pretrained B2)" lines. We generate these two lines by pruning the pretrained SegFormer B2 model weights, with no additional training. Figure~\ref{figure_segformer_pareto_gpu} denotes the model parameters that two example Pareto-optimal configurations $a$ and $b$ correspond to, along with the original B2 model configuration. Surprisingly, model configuration $a$ achieves slightly better mIoU (0.4655 without any retraining and 0.4698 after training) than the original SegFormer B2 ADE model (0.4651). This point has 32 fewer input channels to \textit{Conv2DPred} compared to the B2 model and is 6\% faster on the GPU, so we can start the ADE Pareto Curve from $a$ instead of the full SegFormer ADE B2 model square.

On the GPU, we can save 11\% of execution time with a 1.9\% accuracy loss for SegFormer ADE B2, without any retraining. The pretrained SegFormer City B2 model is more resilient: we can save 11\% of execution time with only a 0.9\% accuracy loss, without any retraining. SegFormer City B2 is trained and executed on larger image sizes (1024 by 1024 vs 512 by 512), so it achieves 1.74$\times$ higher mIoU than SegFormer ADE B2 to begin with. Thus, there is likely more redundancy in these model weights, enabling a flatter tradeoff compared to SegFormer ADE B2.

We plot the original SegFormer B0, B1, and B2 models, labeled as ``ADE Time (Retrained)" and ``City Time (Retrained)" in Figure~\ref{figure_segformer_pareto_gpu}, when executed on the GPU. We can switch between SegFormer models with retrained weights to save 51\% of execution time with a 4.3\% accuracy loss for the ADE B2 model and 45\% of execution time with a 2.5\% accuracy loss for the City B2 model. As expected, switching between pruned models that have been retrained offers a better tradeoff, but the pretrained models are surprisingly resilient: pruning without retraining is competitive up until reducing 15\% of the execution time for the B2 models on the GPU.

\begin{figure}[t]

  \centering
  
  \includegraphics[width=1\linewidth]{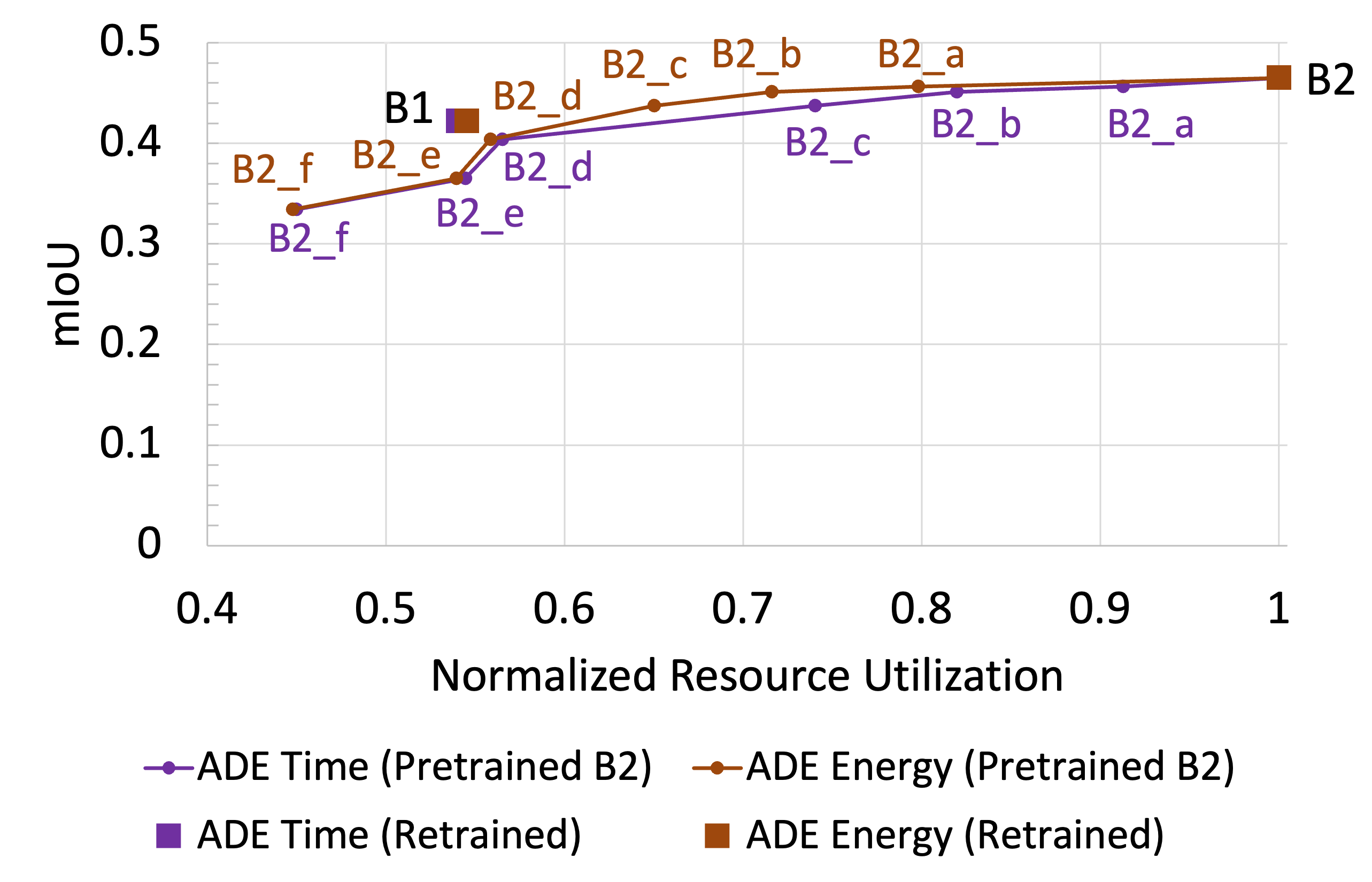}

    % \vspace{-0.1in}
  
  \caption{Tradeoff between execution time, energy, and mIoU (model accuracy) when dynamically pruning pretrained SegFormer B2 models on accelerator E (original retrained model executions shown as squares). The model parameter values for labeled dynamic configurations B2\_a to B2\_f are listed in Table~\ref{table_segformer_pareto_points}. %Dynamic configuration labels have the number of encoder blocks in each of the four encoder stages followed by the input channels for \textit{Conv2DFuse} and \textit{Conv2DPred}; values changed from the original model are underlined. 
  }

  \label{figure_segformer_pareto_accelerator}
  
  % \vspace{-0.2in}
 
\end{figure}

\begin{table}[t]
  {
  \centering

  \begin{tabular}{c|c|c|c} 

    & Number of Encoder & \textit{Conv2DFuse} & \\
    
    Label & Blocks in Stages 0 to 3 & Input Channels & mIoU \\
    
    \midrule
    
    B2 & 3, 4, 6, 3 & 3072 & 0.4651 \\
    B2\_a & 3, 4, 6, 3 & 1920 & 0.4565 \\
    B2\_b & 3, 4, 6, 3 & 1664 & 0.4510 \\
    B2\_c & 2, 4, 6, 3 & 1408 & 0.4374 \\
    B2\_d & 2, 3, 6, 3 & 1024 & 0.4041 \\
    B2\_e & 2, 3, 5, 3 & 896 & 0.3649 \\
    B2\_f & 2, 3, 5, 3 & 512 & 0.3345 \\

  \end{tabular}
    
    \vspace{0.1in}
    
    \caption{SegFormer ADE B2 model execution path configurations, as labeled in Figure~\ref{figure_segformer_pareto_accelerator}. %Label B2 corresponds to the original model execution.
    % \vspace{0.05in}
    }
    \label{table_segformer_pareto_points}
  }

% \vspace{-0.3in}
\end{table}

Figure~\ref{figure_segformer_pareto_accelerator} shows the execution time, energy, and accuracy tradeoff for the various model configurations listed in Table~\ref{table_segformer_pareto_points}, when executed on accelerator E. These points are a subset of the Pareto-optimal configurations found from the GPU experiments. With specialized hardware, we can save 18\% of execution time and 28\% of energy with a 1.4\% accuracy loss for SegFormer ADE B2 without any additional training.

We also plot the original SegFormer B1 and B2 models, labeled as ``ADE Time (Retrained)" and ``ADE Energy (Retrained)" in Figure~\ref{figure_segformer_pareto_accelerator}, and see that with retraining, we can save 55\% of execution time and energy for a 4.3\% accuracy loss. On accelerator E, pruning without retraining reduces execution time and energy by 45\% with the same accuracy loss. Thus, switching between retrained models offers an additional 10\% of execution time and energy savings at the same accuracy for the SegFormer ADE model. One of these approaches can be chosen over the other depending on the range of dynamic resource constraints expected in a real-time system.

\subsection{Swin Model Resilience and Switching}
\label{sec:resiliency_swin}

\begin{figure}[t]

  \centering
  
  \includegraphics[width=1\linewidth]{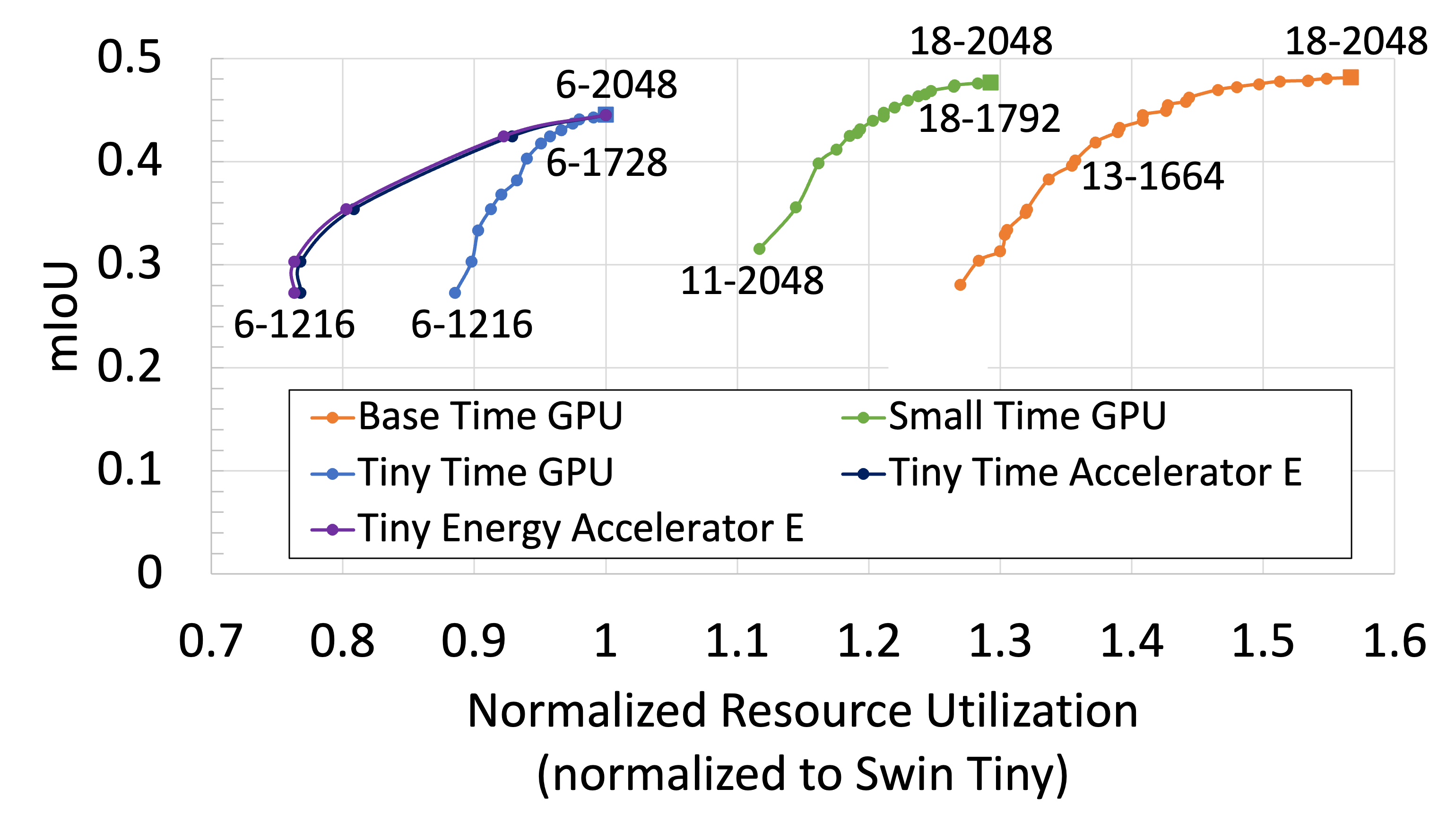}
  \vspace{-0.3in}
  \caption{Tradeoff between execution time, energy and mIoU (model accuracy) when dynamically pruning pretrained the Swin Base, Small, and Tiny models on the A5000 GPU and accelerator E (original retrained model executions on the GPU shown as squares). Dynamic configuration labels list the number of encoder blocks in encoder stage three followed by the number of channels input to \textit{fpn\_bottleneck\_Conv2D}. The resource utilization numbers on the x-axis are normalized to the Swin Tiny execution results.
  }
  
  \label{figure_swin_pareto}
  
  % \vspace{-0.15in}
 
\end{figure}

In Swin-Tiny, almost 90\% of the FLOPs are in convolutions in the decoder. As a result, we vary the input and output channels in these layers along with bypassing encoder blocks, in a similar approach to our SegFormer experiments. Skipping even a few encoder blocks in Swin Tiny leads to a higher drop in model accuracy compared to execution time on the GPU and accelerator E, as shown in Figure~\ref{figure_swin_pareto}, and the marginal accuracy loss quickly outpaces the marginal savings in execution time and energy. 

Swin Small and Swin Base, with $1.35\times$ and $2\times$ as many parameters as Swin Tiny, are slightly more resilient to these optimizations. While all three models have the same number of input channels to the \textit{fpn\_bottleneck\_Conv2D} layer, Swin Small and Base have 18 blocks in encoder stage two compared to only six in Swin Tiny. Thus, we can bypass stage two encoder blocks to achieve a better tradeoff. However, the resilience to pruning is small even for these larger models.

While the tradeoff curve is flatter with accelerator E than with the GPU, we can only save 8\% of execution time and energy with a 2\% accuracy loss for Swin Tiny inference on the accelerator, without any retraining. Thus, unlike with SegFormer, we recommend directly switching between retrained Swin model weights for RDD inference. On the GPU, this approach saves 36\% of execution time for a 3.6\% accuracy drop when switching from Swin Base to Swin Tiny.

While the Swin models require 3.8$\times$ to 4.7$\times$ more FLOPs than SegFormer ADE B2, these larger pretrained models are not more resilient to pruning. The result is because most of the extra FLOPs in Swin are from additional convolutions in the decoder and Swin Tiny has 25\% fewer encoder blocks compared to SegFormer. Thus, while we could bypass encoder blocks to generate a flatter tradeoff curve with SegFormer, the Swin encoder has less redundant information, and the Swin model accuracy is more sensitive to skipping attention layers in the encoder.

\subsection{Once-For-All ResNet-50 Model Switching}
\label{sec:resiliency_ofa}

\begin{figure}[t]

  \centering
  
  \includegraphics[width=1\linewidth]{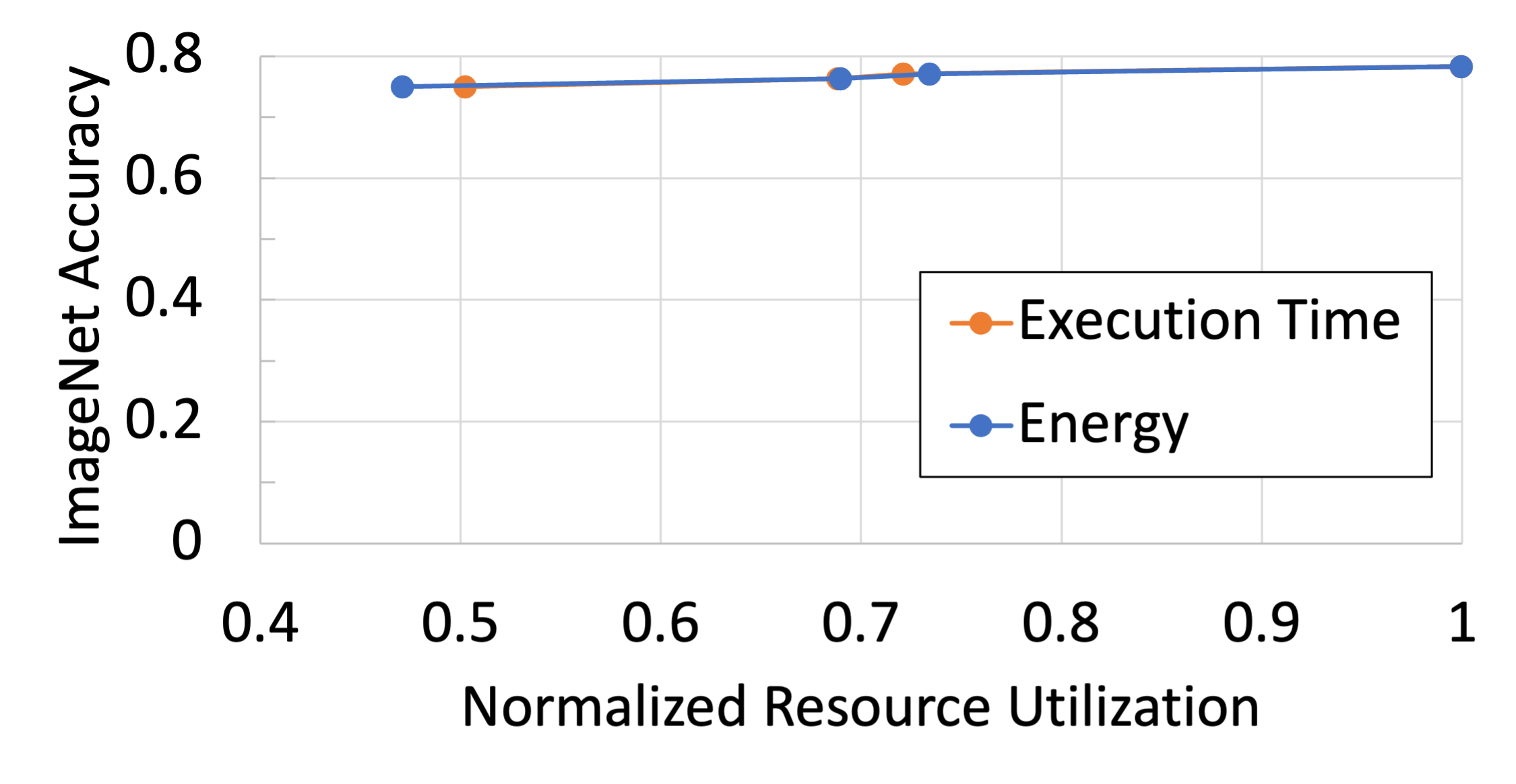}
  
  \vspace{-0.05in}
  
  \caption{Tradeoff between execution time, energy, and model accuracy on the ImageNet dataset, with a 224 by 224 input image size, when dynamically switching between OFA ResNet-50 models on accelerator E.
  }
  
  \label{figure_ofa_pareto}
  
  % \vspace{-0.15in}
 
\end{figure}

Since the ResNet-50 backbone dominates FLOPs for the DETR-based models, we need to scale the work done in this CNN for RDD inference. Fortunately, this is a well-studied area~\cite{cai2019once, jastrzkebski2017residual, clemons2023leaf}, and we leverage switching between OFA ResNet-50 models, which scale model parameters such as the number of layers and convolution kernel sizes. The advantage of the OFA approach is that many model weight subsets are generated within a single training~\cite{cai2019once}. 

We execute various OFA models for ResNet-50 on accelerator E. Figure~\ref{figure_ofa_pareto} shows the effectiveness of switching between these models for dynamic inference on accelerator E: we can save 58\% of the execution time and 53\% of energy with a 3.3\% loss in accuracy. Switching between these models produces a better tradeoff compared to switching between retrained segmentation models on accelerator E, which have comparable execution time and energy savings with a 9\% accuracy loss. This result illustrates that CNNs are more resilient to pruning compared to these segmentation models with the transformer architecture.

\subsection{Principles for identifying alternative model execution paths}
\label{sec:resiliency_principles}

Our results in Sections~\ref{sec:resiliency_seg} to~\ref{sec:resiliency_ofa} highlight general principles to help focus the search for competitive alternative execution paths in vision transformer models. First, we identify that these model executions greatly depend on convolutions, and that we should target convolutional paths in these models instead of pruning attention heads and attention layers as in prior work. Furthermore, our experiments in this section show that convolution weights are less critical to model accuracy compared to attention weights. Second, for modern encoder-decoder architectures, it is advantageous to prune computation in the decoder instead of the encoder. In these attention-free decoders, layers sequentially feed into each other and have one destination, while most encoder layer outputs have two destinations (to the following encoder layers and the decoder). Thus, skipped computation in the decoder can often be propagated backwards through decoder layers to further reduce FLOPs as exemplified with SegFormer in Section~\ref{sec:resiliency_seg}, while this can rarely be done for skipped computation in the encoder. Third, the distribution of FLOPs between the encoder and decoder guides which alternative execution paths will be competitive. When more FLOPs are in the decoder, there is less redundancy in encoder weights, and bypassing encoder layers will lead to higher accuracy losses more quickly, as seen in Section~\ref{sec:resiliency_swin} with pruning Swin Transformer, where 89\% of the FLOPs are in the decoder.

\subsection{RDD Inference Execution}
\label{sec:resiliency_rdd}

% In this section, we have identified alternative model execution paths that bypass computation to meet specific resource targets for our model case studies. During runtime, the inference engine can continually look up which model path meets the resource target input. When enough resources are available, the original model can be executed without any modifications. The engine can optionally output an estimate of the model accuracy.

In this section, we have identified alternative model execution paths that bypass computation to meet specific resource targets for our model case studies. During runtime, the current resource availability is input to the inference engine, which can select the model execution path that matches the resource constraints for that particular inference. Note that this selection is independent of the input image. Each inference task can use a different model configuration that is looked up in real time, as opposed to a pre-selection of a static model to always execute. When enough resources are available, the original model can be executed without any modifications. This means that the average accuracy loss will usually be lower than the percentages reported for particular model configurations in this section, since RDD inference is usually using the full model accuracy.

For more resilient models like SegFormer, we only need to train the original model once and store one set of model weights. During inference, we can use the subset of the weights required for model execution. For other models, we can switch between multiple sets of model weights stored in off-chip DRAM for execution on an accelerator. The OFA approach still requires only one training for this approach with ResNet-50, while switching between the Swin models requires multiple trainings for each model configuration. 

Note that analyzing the resilience of pretrained models provides a floor on the accuracy attainable given a model architecture and a model configuration, while switching between retrained models provides a ceiling. Training techniques such as OFA can be used to improve model accuracy from the floor with less training overheads for RDD inference.

Finally, since we bypass some computation in critical layers for RDD inference, the relative importance of these layers decreases for model configurations with less computation. For example, configuration $B2\_f$ in Figure~\ref{figure_segformer_pareto_accelerator} requires 60\% fewer FLOPs than the full SegFormer ADE B2 model, and \textit{Conv2DFuse} requires less than 25\% of the total FLOPs for this configuration. While the set of critical layers change across model configurations, these layers are still convolutions. For example, 55\% of the total execution time and energy on accelerator E is spent in convolutions for the smaller $B2\_f$ configuration. Thus, we find that accelerators D, E, G are still the best at balancing area, energy, and throughput tradeoffs across all these dynamic model configurations for RDD inference.
%
% This result is not specific to MAGNet; we explore different accelerator parameterizations with the Eyeriss accelerators~\cite{chen2016eyeriss, chen2019eyeriss} for this class of models to find that the same parameters are ideal across our throughput, area, and energy metrics across dynamic model configurations.

%During runtime, the current resource availability is input to select the dynamic model configuration to use for that particular inference. Note that this selection is independent of the input image. Each inference task can use a different model configuration that is looked up in real time, as opposed to a pre-selection of a static model to always execute. This means that the average accuracy loss is usually lower than the smallest and least-accurate dynamic model configuration, since RDD inference is usually using the full model accuracy.
\section{Related Work}
\label{sec:related_work}

% Previous methods for efficient inference largely center on CNNs and BERT models, whereas our focus is on vision transformers.

Static methods such as pruning~\cite{cai2019once, wang2020hat, yang2018netadapt}, knowledge distillation~\cite{sanh2019distilbert, sun2019patient, touvron2021training}, and quantization~\cite{bai2020binarybert, liu2021post, zafrir2019q8bert} reduce model computation but are unable to dynamically adapt to resource constraints during runtime.
Input-dependent methods instead dynamically adjust computation based on input complexity and the state of internal predictions~\cite{xin2020deebert, kaya2019shallow, hu2020triple, graves2016adaptive, liu2020fastbert, figurnov2017spatially, han2021dynamic, dehghani2018universal, teerapittayanon2016branchynet, zhou2020bert, tambe2021edgebert, wang2018skipnet, wang2020dual}, but they reduce inference cost only for inputs that do not require full model execution. Both of these methods can be used with RDD inference to improve efficiency-accuracy tradeoffs.

Few RDD inference methods have been previously explored. One work adds multiple early exit paths in models~\cite{wang2020dual}. Instead of only sequentially skipping layers later in the execution path, we more broadly explore skipping intermediate blocks. Another work augments ResNet and MobileNet with scaled-down versions of blocks of convolutional layers in these models that can be trained and executed instead of the original blocks~\cite{clemons2023leaf}. We instead explore whether the existing model architecture and weights can be leveraged for RDD inference. Compared to both of these prior works, we explore a finer granularity by pruning computation associated with specific channels in critical model layers. In addition, these methods only consider CNNs, while our work is the first to profile the vision transformer applications, focus on vision tasks more complicated than classification, and explore the resilience of pretrained segmentation models compared to retrained models.

% maybe briefly reference these papers here too (already cited in the intro)
% sun2022resource, kim2023aurora
%
% Finally, most hardware accelerators for transformers focus on attention and custom detection components~\cite{li2022divit, wang2022towards, you2022vitcod, zeng2022fpga}, often without support for convolutions. We make the key observation that attention and convolutions are important in these models, and instead leverage accelerators that can efficiently execute both of these layers.

% Few RDD inference methods have been previously explored: one work adds multiple early exit paths in models~\cite{wang2020dual}, while another work augments ResNet and MobileNet with scaled-down versions of blocks of convolutional layers in these models that can be trained and executed instead of the original blocks~\cite{clemons2023leaf}. We instead explore a finer granularity by pruning computation associated with specific channels in critical model layers, and also more broadly explore skipping intermediate layers instead of only sequentially skipping layers later in the execution path. We also explore whether the existing model architecture and weights can be directly leveraged for RDD inference and find that models like SegFormer are fairly resilient to pruning. In addition, these methods mostly consider CNNs, with a few examining BERT, while we instead analyze vision transformers.

\section{Conclusion}

While these vision models are transformers, most of the computation comes from the backbones and task-specific decoder heads. These computations often use convolutions for efficiency, which means that 60\% to 90\% of FLOPs in these applications are in convolutions. On current GPUs, computing CNN FLOPs is very cheap, so these layers take only 20\% to 45\% of the application running time. Thus, accelerating vision transformer applications requires hardware which can handle CNN and transformer computation well. We found that the MAGNet framework worked well for these applications. 

To enable dynamic inference, we found that one needs to bypass both attention-dominated encoder blocks and computation in decoder convolutions to maintain reasonable accuracy. We find that models are more resilient to pruning without additional training when computation is more evenly split between the encoder and the decoder. Surprisingly, even in smaller model configurations, including those with less than half of the original model FLOPs, convolutions still dominate the total computation.

\bibliographystyle{IEEETran}
\bibliography{IEEEexample}

\end{document}